\documentclass[12pt,twoside]{article}
\usepackage[dvips]{graphicx}
\usepackage{latexsym,amssymb,textcomp,wasysym,rotating}
\def\tC{$\cdot$}    
\def\tB{$\bullet$}
\def\tA{$\newmoon$}  

\def\E{\mathbf{E}}

\begin{document}

\title{\bf\Large\hrule height5pt \vskip 4mm
Tests of Machine Intelligence
\vskip 4mm \hrule height2pt}

\author{\\
{\bf Shane Legg}\\[3mm]
\normalsize IDSIA, Galleria 2, Manno-Lugano CH-6928, Switzerland \\
\normalsize \texttt{shane@vetta.org \ \ \ www.vetta.org/shane}
\and \\
{\bf Marcus Hutter}\\[3mm]
\normalsize RSISE$\,$@$\,$ANU and SML$\,$@$\,$NICTA,
\normalsize Canberra, ACT, 0200, Australia \\
\normalsize \texttt{marcus@hutter1.net \ \ \ www.hutter1.net} \\[3ex]
}
\date{December 2007}
\maketitle

\begin{abstract}
Although the definition and measurement of intelligence is clearly of
fundamental importance to the field of artificial intelligence, no
general survey of definitions and tests of machine intelligence
exists.  Indeed few researchers are even aware of alternatives to the
Turing test and its many derivatives.  In this paper we fill this gap
by providing a short survey of the many tests of machine intelligence
that have been proposed.\\
\def\contentsname{\centering\normalsize Contents}
{\parskip=-2.7ex\tableofcontents}
\end{abstract}


\newpage
\section{Introduction}

Despite solid progress on many fronts over the last 50 years,
artificial intelligence is still a very young field with many of its
greatest achievements, and some of its most fundamental problems, yet
to be tackled.  From a theoretical perspective, one of the most
fundamental problems in the field is that the very concept of
intelligence remains rather murky.  This is somewhat true in the
context of humans, but it is especially true when we consider machines
which may have completely different sensors, bodies, cognitive
capacities and live in different environments to ourselves.  What does
``intelligence'' mean for a machine?  Perhaps the first attempt to
answer this question, and certainly the only attempt that most
researchers are aware of, is Alan Turing's famous imitation
game~\cite{Turing:50}.  Turing recognised how difficult it would be to
explicitly define intelligence and thus attempted to sidestep the
issue completely.  Although this was a clever move, it leaves us with
a test of machine intelligence that tells us almost nothing about what
intelligence actually is, and thus is of little use as a foundation,
either theoretical or practical, for our research.

Since then, a few bold researchers have tried to tackle this difficult
problem in a more satisfactory way by proposing various definitions
and tests of machine intelligence.  By and large, these proposals have
been ignored by the community.  Indeed to the best of our knowledge,
no general survey of tests and definitions of intelligence for
machines has ever been published.

We feel that to ignore a question as fundamental as the definition of
machine intelligence is a serious mistake.  In any science, issues
surrounding fundamental definitions and methods of measurement play a
central role and form the foundation on which theoretical advances are
constructed and practical advances are measured.  If we are to truly
advance as a field over the next 50 years, we will need to return to
this most central of problems in order to secure what artificial
intelligence is and what it aims for.  As a first step in this
direction, it is necessary that researchers are at least aware of the
many alternatives to Turing's tests that have been proposed.  In this
paper we hope to partly meet this need by providing the first general
survey of tests and definitions of machine intelligence.

\section{Turing test and derivatives}

The classic approach to determining whether a machine is intelligent
is the so called Turing test \cite{Turing:50} which has been
extensively debated over the last 50 years \cite{Saygin:00}.  Turing
realised how difficult it would be to directly definite intelligence
and thus attempted to side step the issue by setting up his now famous
imitation game: If human judges cannot effectively discriminate
between a computer and a human through teletyped conversation, then we
must conclude that the computer is intelligent.

Though simple and clever, the test has attracted much criticism.
Block and Searle argue that passing the test is not \emph{sufficient}
to establish intelligence \cite{Block:81,Searle:80,Eisner:91}.
Essentially they both argue that a machine could appear to be
intelligent without having any ``real intelligence'', perhaps by using
a very large table of answers to questions.  While such a machine
might be impossible in practice due to the vast size of the table
required, it is not logically impossible.  In which case an
unintelligent machine could, at least in theory, consistently pass the
Turing test.  Some consider this to bring the validity of the test
into question.  In response to these challenges, even more demanding
versions of the Turing test have been proposed such as the Total
Turing test \cite{Harnad:89}, the Truly Total Turing test
\cite{Schweizer:98} and the inverted Turing test \cite{Watt:96}.  Dowe
argues that the Turing test should be extended by ensuring that the
agent has a compressed representation of the domain area, thus ruling
out look-up table counter arguments~\cite{Dowe:98}.  Of course these
attacks on the Turing test can be applied to any test of intelligence
that considers only a system's external behaviour, that is, most
intelligence tests.

A more common criticism is that passing the Turing test is not
\emph{necessary} to establish intelligence.  Usually this argument is
based on the fact that the test requires a machine to have a highly
detailed model of human knowledge and patterns of thought, making it a
test of humanness rather than intelligence \cite{French:90,Ford:98}.
Indeed even small things like pretending to be \emph{unable} to
perform complex arithmetic quickly and faking human typing errors
become important, something which clearly goes against the purpose of
the test.

The Turing test has other problems as well.  Current AI systems are a
long way from being able to pass an unrestricted Turing test.  From a
practical point of view this means that the full Turing test is unable
to offer much guidance to our work.  Indeed, even though the Turing
test is the most famous test of machine intelligence, almost no
current research in artificial intelligence is specifically directed
toward being able to pass it.  Unfortunately, simply restricting the
domain of conversation in the Turing test to make the test easier, as
is done in the Loebner competition \cite{Loebner:90}, is not
sufficient.  With restricted conversation possibilities the most
successful Loebner entrants are even more focused on faking human
fallibility, rather than anything resembling intelligence
\cite{Hutchens:96}.  Perhaps a better alternative then is to test
whether a machine can imitate a child (see for example the tests
described in Sections~\ref{sec:linguistic} and~\ref{sec:toddler}).
Finally, the Turing test returns different results depending on who
the human judges are.  Its unreliability has in some cases lead to
clearly unintelligent machines being classified as human, and at least
one instance of a human actually failing a Turing test.  When queried
about the latter, one of the judges explained that ``no human being
would have that amount of knowledge about
Shakespeare''\cite{Shieber:94}.

\section{Compression tests}

Mahoney has proposed a particularly simple solution to the binary pass
or fail problem with the Turing test: Replace the Turing test with a
text compression test \cite{Mahoney:99}.  In essence this is somewhat
similar to a ``Cloze test'' where an individual's comprehension and
knowledge in a domain is estimated by having them guess missing words
from a passage of text.

While simple text compression can be performed with symbol
frequencies, the resulting compression is relatively poor.  By using
more complex models that capture higher level features such as aspects
of grammar, the best compressors are able to compress text to about
1.5 bits per character for English.  However humans, which can also
make use of general world knowledge, the logical structure of the
argument etc., are able to reduce this down to about 1 bit per
character.  Thus the compression statistic provides an easily computed
measure of how complete a machine's model of language, reasoning and
domain knowledge are, relative to a human.

To see the connection to the Turing test, consider a compression test
based on a very large corpus of dialogue.  If a compressor could
perform extremely well on such a test, this is mathematically
equivalent to being able to determine which sentences are probable at
a given point in a dialogue, and which are not (for the equivalence of
compression and prediction see~\cite{Bell:90}).  Thus, as failing a
Turing test occurs when a machine (or person!) generates a sentence
which would be improbable for a human, extremely good performance on
dialogue compression implies the ability to pass a Turing test.

A recent development in this area is the Hutter
Prize~\cite{Hutter:06hprize}.  In this test the corpus is a 100 MB
extract from Wikipedia.  The idea is that this should represent a
reasonable sample of world knowledge and thus any compressor that can
perform very well on this test must have a good model of not just
English, but also world knowledge in general.

One criticism of compression tests is that it is not clear whether a
powerful compressor would easily translate into a general purpose
artificial intelligence.

\section{Linguistic complexity}\label{sec:linguistic}

A more linguistic approach is taken by the HAL project at the company
Artificial Intelligence NV \cite{Treister:01}.  They propose to
measure a system's level of conversational ability by using techniques
developed to measure the linguistic ability of children.  These
methods examine things such as vocabulary size, length of utterances,
response types, syntactic complexity and so on.  This would allow
systems to be ``\ldots assigned an age or a maturity level beside
their binary Turing test assessment of `intelligent' or `not
intelligent'~''\cite{Treister:00}.  As they consider communication to
be the basis of intelligence, and the Turing test to be a valid test
of machine intelligence, in their view the best way to develop
intelligence is to retrace the way in which human linguistic
development occurs.  Although they do not explicitly refer to their
linguistic measure as a test of intelligence, because it measures
progress towards what they consider to be a valid intelligence test,
it acts as one.

\section{Multiple cognitive abilities}\label{sec:toddler}

A broader developmental approach is being taken by IBM's Joshua Blue
project \cite{Alvarado:02}.  In this project they measure the
performance of their system by considering a broad range of
linguistic, social, association and learning tests.  Their goal is
to first pass what they call a ``toddler Turing test'', that is, to
develop an AI system that can pass as a young child in a similar
setup to the Turing test.  As yet, this test is not fully specified.

Another company pursuing a similar developmental approach based on
measuring system performance through a broad range of cognitive tests
is the a2i2 project at Adaptive AI \cite{Voss:05}.  Rather than
toddler level intelligence, their current goal to is work toward a
level of cognitive performance similar to that of a small mammal.  The
idea being that even a small mammal has many of the key cognitive
abilities required for human level intelligence working together in an
integrated way.  While this might be useful to guide the development
of moderate intelligence, it is unknown whether it will scale to
higher levels of intelligence.  The specific tests being used have not
been published.

\section{Competitive games}

The Turing Ratio method of Masum et al.\ has more emphasis on tasks
and games rather than cognitive tests.  They propose that ``\ldots
doing well at a broad range of tasks is an empirical definition of
`intelligence'."\cite{Masum:02} To quantify this they seek to identify
tasks that measure important abilities, admit a series of strategies
that are qualitatively different, and are reproducible and relevant
over an extended period of time.  They suggest a system of measuring
performance through pairwise comparisons between AI systems that is
similar to that used to rate players in the international chess rating
system.  The key difficulty however, which the authors acknowledge is
an open challenge, is to work out what these tasks should be, and to
quantify just how broad, important and relevant each is.  In our view
these are some of the most central problems that must be solved when
attempting to construct an intelligence test and thus this approach is
incomplete in its current state.

\section{Collection of psychometric tests}

An approach called Psychometric AI tries to address the problem of
what to test for in a pragmatic way.  In the view of Bringsjord and
Schimanski, ``Some agent is intelligent if and only if it excels at
all established, validated tests of [human]
intelligence.''\cite{Bringsjord:03} They later broaden this to also
include ``tests of artistic and literary creativity, mechanical
ability, and so on.''  With this as their goal, their research is
focused on building robots that can perform well on standard
psychometric tests designed for humans, such as the Wechsler Adult
Intelligent Scale and Raven Progressive Matrices.

As effective as these tests are for humans, they seem inadequate for
measuring machine intelligence as they are highly anthropocentric and
embody basic assumptions about the test subject that are likely to be
violated by computers.  For example, consider the fundamental
assumption that the test subject is not simply a collection of
specialised algorithms designed only for answering common IQ test
questions.  While this is obviously true of a human, or even an ape,
it may not be true of a computer.  The computer could be nothing more
than a collection of specific algorithms designed to identify patterns
in shapes, predict number sequences, write poems on a given subject or
solve verbal analogy problems --- all things that AI researchers have
worked on.  Such a machine might be able to obtain a respectable IQ
score \cite{Sanghi:03}, even though outside of these specific test
problems it would be next to useless.  If we try to correct for these
limitations by expanding beyond standard tests, as Bringsjord and
Schimanski seem to suggest, this once again opens up the difficulty of
exactly what, and what not, to test for.  Psychometric AI, at least as
it is currently formulated, only partially addresses this central
question.

\section{Smith's test}

The basic structure of Smith's test is that an agent faces a series of
problems that are generated by an algorithm~\cite{Smith:06}.  In each
iteration the agent must try to produce the correct response to the
problem that it has been given.  The problem generator then responds
with a score of how good the agent's answer was.  If the agent so
desires it can submit another answer to the same problem.  At some
point the agent requests to the problem generator to move onto the
next problem and the score that the agent received for its last answer
to the current problem is then added to its cumulative score.  Each
interaction cycle counts as one time step and the agent's intelligence
is then its total cumulative score considered as a function of time.
In order to keep things feasible, the problems must all be in P, i.e.\
the solution must be verifiable in polynomial time.

We have two main criticisms of Smith's definition.  Firstly, while for
practical reasons it might make sense to restrict problems to be in P,
we do not see why this practical restriction should be a part of the
very definition of intelligence as Smith suggests.  If some
breakthrough meant that agents could solve difficult problems in not
just P but sometimes in NP as well, then surely these new agents would
be more intelligent?

Secondly, while the definition is somewhat formally defined, it still
leaves open the important question of what exactly the tests should
be.  Smith suggests that researchers should dream up tests and then
contribute them to some common pool of tests.  As such, this is not a
fully specified test.

\section{C-Test}

One perspective among psychologists who support the $g$-factor view of
intelligence, is that intelligence is ``the ability to deal with
complexity''\cite{Gottfredson:97}.  Thus in a test of intelligence the
most difficult questions are the ones that are the most complex
because these will, by definition, require the most intelligence to
solve.  It follows then that if we could formally define and measure
the complexity of test problems we could construct a formal test of
intelligence.  The possibility of doing this was perhaps first
suggested by the complexity theorist Chaitin \cite{Chaitin:82}.  While
this path requires numerous difficulties to be dealt with, we believe
that it is the most natural and offers many advantages: It is formally
motivated, precisely defined and potentially could be used to measure
the performance of both computers and biological systems on the same
scale without the problem of bias towards any particular species or
culture.

One intelligence test that is based on formal complexity theory is the
C-Test from Hern\'{a}ndez~\cite{Hernandez:00cmi,Hernandez:98fdi}.
This test consists of a number of sequence prediction and abduction
problems similar to those that appear in many standard IQ tests.
Similar to standard IQ tests, the C-Test always ensures that each
question has an unambiguous answer in the sense that there is always
one hypothesis that is consistent with the observed pattern that has
significantly lower complexity than the alternatives.  The key
difference to sequence problems that appear in standard intelligence
tests is that the questions are based on a formally expressed measure
of complexity, namely Levin's computable $Kt$
complexity \cite{Levin:73search} (rather than Kolmogorov's
incomputable complexity \cite{Li:97}) to get a practical test.  In
order to retain the invariance property of Kolmogorov complexity,
Levin complexity requires the additional assumption that the universal
Turing machines are able to simulate each other in linear time.

The test has been successfully applied to humans with intuitively
reasonable results \cite{Hernandez:98fdi,Hernandez:00btt}.  As far as
we know, this is the only formal definition of intelligence that has
so far produced a usable test of intelligence.

One criticism of the C-Test and Smith's tests is that the way
intelligence is measured is essentially static, that is, the
environments are passive.  We believe that dynamic testing in active
environments is a better measure of a system's intelligence.  To put
this argument another way: Succeeding in the real world requires you
to be more than an insightful spectator!  One must carefully choose
actions knowing that these may affect the future.

\section{Universal intelligence}

Another complexity based test is the
\emph{universal intelligence} test~\cite{Legg:06ior}.  Unlike the
C-Test and Smith's test, universal intelligence tests the performance
of an agent in a fully interactive environment.  This is done by using
the reinforcement learning framework in which the agent sends
its \emph{actions} to the environment and receives
\emph{observations} and \emph{rewards} back.  The agent tries to
maximise the amount of reward it receives by learning about the
structure of the environment and the goals it needs to accomplish in
order to receive rewards.

Formally, the process of interaction produces an increasing history
$o_1 r_1 a_1 o_2 r_2 a_2 o_3 r_3 a_3 o_4 \ldots$ of observations
$o$, rewards $r\geq 0$, and actions $a$.  The agent is simply a
function, denoted by $\pi$, which is a probability measure over
actions conditioned on the current history, for example, $\pi( a_3 |
o_1 r_1 a_1 o_2 r_2 )$.  The environment, denoted $\mu$, is
similarly defined: $\mu( o_k r_k | o_1 r_1 a_1 o_2 r_2 a_2 \ldots
o_{k-1} r_{k-1} a_{k-1} )$.
The performance of agent $\pi$ in environment $\mu$ can be measured
by its total expected reward $V^\pi_\mu := \E[\sum_{i=1}^{\infty}
r_i|\mu,\pi]$, called value. The largest interesting class of
environments is the class $E$ of all computable probability
distributions $\mu$. For technical reasons, the values are assumed
to be bounded by some constant $c$.

To get a single performance measure $V^\pi_\mu$ is averaged over all
$\mu\in E$. As there are an infinite number of environments, with no
bound on their complexity, it is impossible to take the expected
value with respect to a uniform distribution --- some environments
must be weighted more heavily than others. Considering the agent's
perspective on the problem, it is the same as asking: Given several
different hypotheses which are consistent with the observations,
which hypothesis should be considered the most likely?  This is a
fundamental problem in inductive inference for which the standard
solution is to invoke Occam's razor: \emph{Given multiple hypotheses
which are consistent with the data, simpler ones should be
preferred.} As this is generally considered the most intelligent
thing to do, one should test agents in such a way that they are, at
least on average, rewarded for correctly applying Occam's razor.
This means that the a priori distribution over environments should
be weighted towards simpler environments.

As each environment $\mu$ is described by a computable measure,
their complexity can be measured with Kolmogorov complexity
$K(\mu)$, which is simply the length of the shortest program
that computes $\mu$ \cite{Li:97}.
The right a priori weight for $\mu$ is $2^{-K(\mu)}$.
We can now define the \emph{universal intelligence} of an agent
$\pi$ to simply be its expected performance,
\[
\Upsilon (\pi) := \sum_{\mu \in E} 2^{-K(\mu)} V^\pi_\mu.
\]
By construction, universal intelligence measures the general ability of
an agent to perform well in a very wide range of environments, similar
to the essence of many informal definitions of
intelligence~\cite{Legg:07idefs}.  The definition places no
restrictions on the internal workings of the agent; it only requires
that the agent is capable of generating output and receiving input
which includes a reward signal.
If we wish to bias the test to reflect world knowledge then we can
condition the complexity measure.  For example, use $K(\mu|D)$ where
$D$ is some set of background knowledge such as Wikipedia.

By considering $V^\pi_\mu$ for a number of basic environments, such as
small MDPs, and agents with simple but very general optimisation
strategies, it is clear that $\Upsilon$ correctly orders the relative
intelligence of these agents in a natural way.  A very high value of
$\Upsilon$ would imply that an agent is able to perform well in many
environments.  The maximal agent with respect to $\Upsilon$ is the
theoretical AIXI agent which has been shown to have many strong
optimality properties \cite{Hutter:04uaibook}.  These results confirm
that agents with high universal intelligence are indeed very
powerful and adaptable.
Universal intelligence spans simple adaptive agents right up to
super intelligent agents like AIXI.  The test is completely formally
specified in terms of fundamental concepts such as universal Turing
computation and complexity and thus is not anthropocentric.

A test based on $\Upsilon$ would evaluate the performance of an agent
on a large sample of simulated environments, and then combine the
agent's performance in each environment into an overall intelligence
value.
The key challenge that needs to be dealt with is to find a suitable
replacement for the incomputable Kolmogorov complexity function,
possibly Levin's $Kt$ complexity \cite{Levin:73search}, as is done
by the C-Test.

\section{Summary}

We end this survey with a comparison of the various tests considered.
Table~\ref{table:mitests} rates each test according to the properties
described below. Although we have attempted to be as fair as possible,
some of the scores we give on this table will naturally be
debatable. Nevertheless, we hope that it provides a rough overview of
the relative strengths and weaknesses of the proposals.

\emph{Valid}: A test of intelligence should capture intelligence and
not some related quantity.  \emph{Informative}: The result should be a
scalar value, or perhaps a vector.  \emph{Wide range}: A test should
cover low levels of intelligence up to super intelligence.
\emph{General}: Ideally we would like to have a very general test that
could be applied to everything from a fly to a machine learning
algorithm.  \emph{Dynamic}: A test should directly take into account
the ability to learn and adapt over time.  \emph{Unbiased}: A test
should not be biased towards any particular culture, species, etc.
\emph{Fundamental}: We do not want a test that needs to be changed
from time to time due to changing technology and knowledge.
\emph{Formal}: The test should be precisely defined, ideally using
mathematics.  \emph{Objective}: The test should not appeal to
subjective assessments such as the opinions of human judges.
\emph{Fully Defined}: Has the test been fully defined, or are parts
still unspecified?  \emph{Universal}: Is the test universal, or is it
anthropocentric?  \emph{Practical}: A test should be able to be
performed quickly and automatically.  \emph{Test vs. Def}: Finally we
note whether the proposal is more of a test, more of a definition, or
something in between.

\begin{table}
\caption{\label{table:mitests} In the table \tA\ means ``yes'',
  \tB\ means ``debatable'', \tC\ means ``no'', and ? means unknown.
  When something is rated as unknown that is usually because the test
  in question is not sufficiently specified.}
\begin{small}
\begin{tabular}{l|c|c|c|c|c|c|c|c|c|c|c|c|c|}
 \\ \\
\multicolumn{1}{l|}{Intelligence Test} \\
&
\multicolumn{1}{c}{\begin{rotate}{45}Valid\end{rotate}} &
\multicolumn{1}{c}{\begin{rotate}{45}Informative\end{rotate}} &
\multicolumn{1}{c}{\begin{rotate}{45}Wide Range\end{rotate}} &
\multicolumn{1}{c}{\begin{rotate}{45}General\end{rotate}} &
\multicolumn{1}{c}{\begin{rotate}{45}Dynamic\end{rotate}} &
\multicolumn{1}{c}{\begin{rotate}{45}Unbiased\end{rotate}} &
\multicolumn{1}{c}{\begin{rotate}{45}Fundamental\end{rotate}} &
\multicolumn{1}{c}{\begin{rotate}{45}Formal\end{rotate}} &
\multicolumn{1}{c}{\begin{rotate}{45}Objective\end{rotate}} &
\multicolumn{1}{c}{\begin{rotate}{45}Fully Defined\end{rotate}} &
\multicolumn{1}{c}{\begin{rotate}{45}Universal\end{rotate}} &
\multicolumn{1}{c}{\begin{rotate}{45}Practical\end{rotate}} &
\multicolumn{1}{c}{\begin{rotate}{45}Test vs. Def.\end{rotate}} \\
\hline
Turing Test            &\tB & \tC &\tC &\tC &\tB &\tC &\tC &\tC &\tC &\tB &\tC &\tB & T \\
Total Turing Test      &\tB & \tC &\tC &\tC &\tB &\tC &\tC &\tC &\tC &\tB &\tC &\tC & T \\
Inverted Turing Test   &\tB & \tB &\tC &\tC &\tB &\tC &\tC &\tC &\tC &\tB &\tC &\tB & T \\
Toddler Turing Test    &\tB & \tC &\tC &\tC &\tB &\tC &\tC &\tC &\tC &\tC &\tC &\tB & T \\
Linguistic Complexity  &\tB & \tA &\tB &\tC &\tC &\tC &\tC &\tB &\tB &\tC &\tB &\tB & T \\
Text Compression Test  &\tB & \tA &\tA &\tB &\tC &\tB &\tB &\tA &\tA &\tA &\tB &\tA & T \\
Turing Ratio           &\tB & \tA &\tA &\tA & ?  & ?  & ?  & ?  & ?  &\tC & ?  & ?  & T/D \\
Psychometric AI        &\tA & \tA &\tB &\tA & ?  &\tB &\tC &\tB &\tB &\tB &\tC &\tB & T/D \\
Smith's Test           &\tB & \tA &\tA &\tB &\tC & ?  &\tA &\tA &\tA &\tC & ?  &\tB & T/D \\
C-Test                 &\tB & \tA &\tA &\tB &\tC &\tA &\tA &\tA &\tA &\tA &\tA &\tA & T/D \\
Universal Intelligence &\tA & \tA &\tA &\tA &\tA &\tA &\tA &\tA &\tA &\tA &\tA &\tC & D \\
\cline{2-14}
\end{tabular}
\end{small}
\end{table}

\subsubsection*{Acknowledgements.}

This work was supported by the Swiss NSF grant 200020-107616.

\bibliographystyle{plain}

\begin{small}

\end{small}

\end{document}